\documentclass[10pt,twocolumn,letterpaper]{article}

\usepackage{iccv}
\usepackage{times}
\usepackage{epsfig}
\usepackage{graphicx}
\usepackage{amsmath}
\usepackage{amssymb}
\usepackage{bbm}
\usepackage{amsfonts}
\usepackage{mathrsfs}
\usepackage{multirow}
\usepackage{enumitem}
\usepackage{algpseudocode}
\usepackage{booktabs}

\usepackage[pagebackref=true,breaklinks=true,letterpaper=true,colorlinks,bookmarks=false]{hyperref}

\iccvfinalcopy % *** Uncomment this line for the final submission

\def\iccvPaperID{***} % *** Enter the CVPR Paper ID here

\begin{document}

%%%%%%%%% TITLE
\title{Slow~Perception:~Let's~Perceive~Geometric~Figures~Step-by-step}

% \author{**** \\ ****}
\author{Haoran Wei{$^{1}$}\thanks{Equal contribution},~~Youyang Yin{$^{2*}$},~~ Yumeng Li{$^{2}$},~~ Jia Wang{$^{1}$}, \\
Liang Zhao{$^{1}$},~~Jianjian Sun{$^{1}$},~~Zheng Ge{$^{1}$},~~Xiangyu Zhang{$^{1}$},~~Daxin Jiang{$^{1}$} \\
{$^{1}$}StepFun~~~~~ {$^{2}$}Beihang University \\
\small{{\url{https://github.com/Ucas-HaoranWei/Slow-Perception}}}
}


\maketitle
\thispagestyle{empty}

\begin{abstract}
% Geometry parsing presents a formidable challenge for current Vision-Language Models (VLMs). Despite these models' advanced perception capabilities, 
% While modern advanced Large Vision Language Models (LVLMs) excel at instantaneous vision recognition, they struggle with fine-grained perception tasks, e.g., geometry parsing, due to the intricate element relationships within geometric shapes. 
Recently, ``visual o1" began to enter people’s vision, with expectations that this slow-thinking design can solve visual reasoning tasks, especially geometric math problems. However, the reality is that current LVLMs (Large Vision Language Models) can hardly even accurately copy a geometric figure, let alone truly understand the complex inherent logic and spatial relationships within geometric shapes. We believe accurate copying (strong perception) is the first step to visual o1. Accordingly, we introduce the concept of “slow perception” (SP), which guides the model to gradually perceive basic point-line combinations, as our humans, reconstruct complex geometric structures progressively. There are two-fold stages in SP: a) perception decomposition. Perception is not instantaneous. In this stage, complex geometric figures are broken down into basic simple units to unify geometry representation. b) perception flow, which acknowledges that accurately tracing a line is not an easy task. This stage aims to avoid ``long visual jumps'' in regressing line segments by using a proposed ``perceptual ruler'' to trace each line stroke-by-stroke.
Surprisingly, such a human-like perception manner enjoys an inference time scaling law—the slower, the better. Researchers strive to speed up the model’s perception in the past, but we slow it down again, allowing the model to read the image step-by-step and carefully.  

% All our data, benchmarks, and code will be open-sourced. 

%  Moreover, there is no definitive conclusion on whether careful reading of an image requires thinking or not.

\end{abstract}

\section{Introduction}

\begin{figure}[t]
\centering
\includegraphics[width=8.3cm]{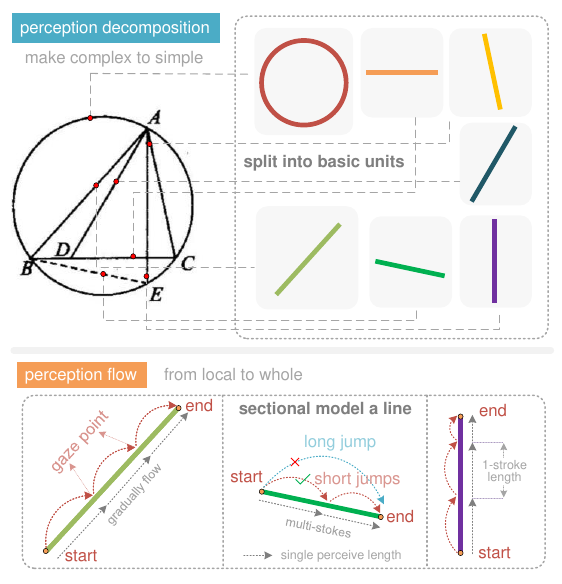}\\
\caption{Slow perception enjoys two stages: 1) Perception decomposition. A geometric shape is decomposed into basic visual units, such as circles and line segments, thereby unifying the fundamental representational form of diverse geometric figures. 2) Perception flow.  Using the same modeling approach (predicting the endpoint based on the starting point) for line segments of different lengths is unreasonable.  We employ a sectional copying method to express each line segment with a perceptual ruler.}
\label{fig:1} 
\end{figure}

Geometric figure parsing, entailing the conversion of geometric shapes in 2D images into editable, is a significant task in computer vision, which enjoys promising academic and industrial values. In the realm of research, geometric figure perception has the potential to prompt the mathematical visual reasoning field~\cite{yue2024mmmu,chen2021geoqa,masry2022chartqa,trinh2024solving}. Meanwhile, in applied domains, it also holds landing prospects in education, architecture, and other fields. However, geometry parsing is not easy due to the spatial relationships and dependencies among geometric units. To our knowledge, there are no effective solutions, pretrain data, or valid benchmarks so far, which further hinders the development of this field.

Over the last few years, when detection algorithms were particularly popular~\cite{ren2015faster,redmon2017yolo9000,lin2017focal,law2018cornernet,wei2022corner,wei2022humanliker,wei2022cornerformer}, utilizing detection models for geometric parsing is a considered approach~\cite{kalleli2024historical}. However, different from natural objects~\cite{COCO}, geometric shapes inherently possess element relations. For example, in $\angle ABC$ as shown in Figure~\ref{fig:1}, the sides $AB$ and $BC$ converge at the common vertex $B$, whereas object detection methods predict object targets independently (in parallel). As a result, the output results of line $AB$ and $BC$ may yield inconsistent coordinates for point $B$. This decoupling prediction manner is why traditional detection struggles with geometric figure parsing tasks.

\begin{figure}[t]
\centering
\includegraphics[width=8.1cm]{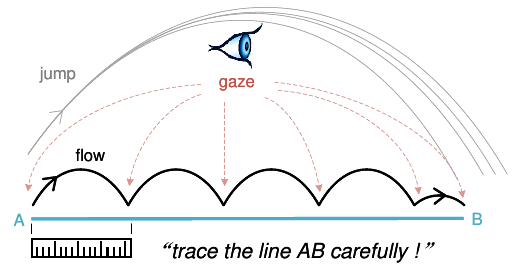}\\
\caption{When humans trace a line, it is typically a slow perception process. Rather than sketching the line, especially a long line, in one stroke (long range ``jump"), humans commonly draw a line with ``multiple short strokes" for high precision. Our ``slow perception" algorithm is designed based on this to mimic the gradual human process of discerning geometric figures.}
\label{fig:2} 
\end{figure}

In the past two years, LVLMs~\cite{GPT4, Qwen-VL, wei2023vary, wei2024small_varytoy, zhao2023chatspot} have demonstrated exceptional capabilities in image description~\cite{COCO} and visual question-answering~\cite{TextVQA,DocVQA} tasks. More importantly, for the geometry parsing task, the next token prediction modeling approach ensures that subsequent points can reference the coordinates of previously generated ones, thereby guaranteeing the closure of the output geometric shape, leading us to hope LVLMs could solve geometry parsing problems.  However, when we attempt to use state-of-the-art models~\cite{belouadi2024detikzify,wei2024general,Qwen-VL,GPT4,llava}, to generate code for de-rendering geometric shapes, we find all of them, even like GPT-4o~\cite{GPT4} and Claude3.5~\cite{AnthropicClaude}, fail to demonstrate this capability. For us humans, copy a geometric shape seems to be a straightforward task that even an elementary school student can perform well with just a ruler. This compels us to pay close attention to this task. A natural question arises: \textit{where do current LVLM modeling paradigms fall short?}

Imagine if we are to trace a geometric shape manually. We never accomplish this in one fell swoop (in one stoke). Instead, the typical methodology involves: 1)  disassembling complex geometric shapes into small units (aiming to ``make complex to simple''); and 2) drawing each visual part stroke-by-stroke (``from local to a whole''), as shown in Figure~\ref{fig:2}. We claim it is the answer to the above question, and following this, we propose the ``slow perception'' (SP) concept to guide the model to do such a task as humans.

Specifically, regardless of how complex a geometric shape may be, it can always be decomposed into the most basic combinations of points and lines. This allows for a unified geometric representation of all shapes. For instance, don't need to care about what polygon a geometric shape is, the model only needs to predict each line segments that compose it in a certain order. This we call the perception decomposition (1-order slow-down). However, modeling a line segment is not as simple as just considering it as paired endpoints. This definition faces two problems: 1) The number of tokens that represent the line segment is fewer than those representing the endpoints, leading to insufficient optimization of the line segment (the relation between points). This can result in accurate point prediction but chaotic connections between points. 2) The computational cost for predicting long and short lines is the same, which contradicts our intuitive perception. Inspired by the ruler tool and eye movement process humans use when copying geometric line segments (Figure~\ref{fig:2}), we propose the perception flow (2-order perception slow-down), which employs a segmented tracing method to represent each line. Specifically, each line segment can be represented as: ``start point $\to$ gaze point 1 $\to$ gaze point 2 $\to$ $\cdots$ $\to$ gaze point $n$ $\to$ endpoint''. The value of $n$ is related to both the length of the target line segment and the preset ``perceptual ruler".

Most importantly, along with the concept and modeling approach of slow perception, we provide a method for rendering geometric shapes to scale up the dataset. We have constructed a total of 200,000 synthetic data samples for training a model. Additionally, we collect manually annotated 480 real-world geometric figures from middle school exam paper scenarios, with 120 for validation and 360 as a test set. We will open-source all data and codebase to promote community development in such a field.

Experimentally, slow perception can improve the F1-score by 6\%. We also discover two interesting conclusions: 1) The perception flow method for line prediction consistently improves accuracy. Even when the perceptual ruler is set to a relatively large value, it can solidly enhance performance. 2) Slow perception exhibits an \textbf{inference time scaling law}: as the perceptual ruler decreases in length, the computational cost for predicting each line segment increases, leading to longer inference times, which gradually improves the geometric parsing performance. 

In summary, geometric shapes are human abstractions of natural vision objects. Thereby, we believe our findings in the geometry parsing task will provide insights for other research areas of computer vision as well.

% For those models specifically trained on geometry code, such as Detikzy~\cite{belouadi2024detikzify} and GOT-OCR2.0~\cite{wei2024general}, their generalization ability is also unacceptable. Despite potentially performing well on specific datasets, these models struggle when encountering the diversity and complexity of real-world geometric parsing tasks. This indicates that while current technological advancements offer new possibilities for geometric parsing, significant challenges remain in developing truly reliable and efficient solutions.

\begin{figure*}[!t]%[h!]
	\centering
	\includegraphics[width=17.2cm]{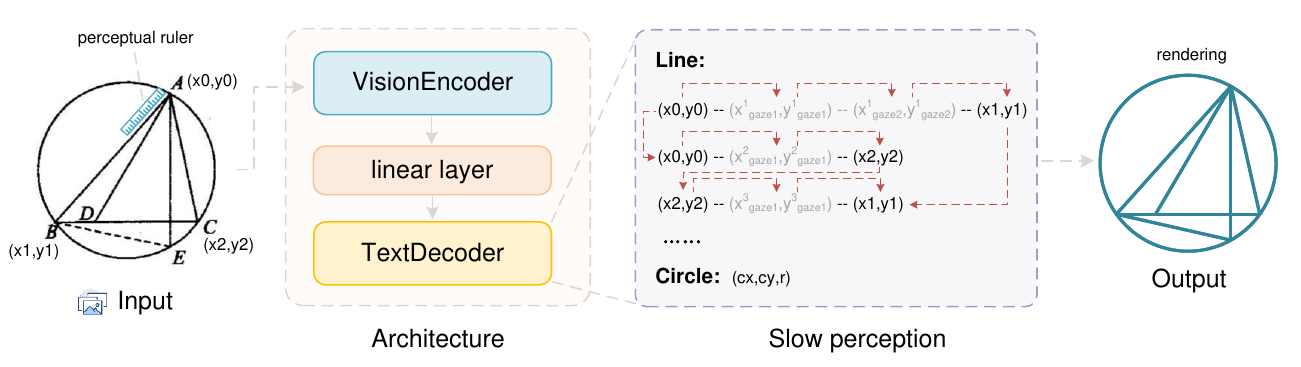}
	%\caption{pic1}
	%\vspace{-2mm}
	\caption{The framework of slow perception. Our approach is adaptable to the most popular LVLM frameworks. According to the next-token serialized prediction, predicted subsequent geometric points can reference the coordinates of preceding points to achieve closed shapes more easily. We establish a perceptual ruler as the upper limit for single-step distance prediction.}
	\label{Fig3}
\end{figure*}

\section{Related Works}
\subsection{Object Detection as Vision Perception}
Object detection~\cite{girshick2014rich,girshick2015fast,ren2015faster,redmon2016you,redmon2017yolo9000} is one of the hottest research topics in computer vision, which can be broadly categorized into two-stage~\cite{ren2015faster} and one-stage~\cite{redmon2016you}. Previously, it was believed that two-stage methods offered higher accuracy while one-stage methods were faster. Later, with the further development of foundational models, algorithmic engineering, and transformer~\cite{carion2020end} networks, one-stage models have become both powerful and efficient. In recent years, the prevailing trend in detection algorithms seems to have been dominated by the one-stage type. 

For the geometric parsing task, using object detection algorithms does not seem to make sense. This is because independently detecting each geometric visual component cannot guarantee the whole geometric closure. For instance, point $A$ often serves as the endpoint of multiple different line segments, and the parallel prediction of each line cannot ensure the consistency of this point, even if the error is minimal. Furthermore, from RCNN~\cite{girshick2014rich} to Faster RCNN~\cite{ren2015faster}, and then to the YOLO series~\cite{redmon2016you,redmon2017yolo9000}, this is a trend towards increasingly faster perception. However, we can't help but question: is faster perception always better? is perception purely an optimization problem? do we annotate objects in dense areas, small objects, or extremely large objects at the same speed?

\subsection{LVLM for Vision Perception}

Recently, research on large vision-language models (LVLMs)~\cite{llava,Qwen-VL,ye2023mplug,chen2024far_intervl1.5} has been on the rise, and these models have demonstrated state-of-the-art performance in various visual perception tasks, such as OCR~\cite{TextVQA,wei2024general,liu2024focus_fox,chen2024onechart} and grounding~\cite{zhao2023chatspot,yu2023merlin}. After more than a year of development, the framework of these LVLMs has become quite convergent. Specifically, new models often adopt an ``encoder-perceiver-decoder" architecture and utilize a training approach similar to large language models (LLMs), primarily involving pretraining followed by supervised fine-tuning (SFT). It is worth noting that the powerful visual knowledge (open-set universal object recognition capability) of LVLMs has also left a deep impression on people, leading us to have very high expectations for LVLMs.

However, some works like  BlindTest~\cite{rahmanzadehgervi2024vision} show that LVLMs don’t seem to understand images truly; in other words, the models look at images too superficially. This cursory glance manner of reading makes it difficult to capture details, logic, and spatial relationships within the image. Some works have attempted to enhance VLM capabilities using a chain of thought~\cite{wei2022chain} approach. What’s puzzling is: does perceiving an object multiple times; or only reading an image carefully, require thinking?

\section{Methodology}

\subsection{Architecture}

As shown in Figure~\ref{Fig3}, we chose the classic LVLM framework for experiments to verify the efficiency of slow perception. It usually consists of a vision encoder preceding an LLM decoder, with a simple linear layer in between for channel mapping. Specifically, we use GOT-OCR2.0~\cite{wei2024general} as the primary experimental model due to its iterative efficiency. Additionally, we utilize other classic LVLMs, e.g., Qwen2-VL~\cite{wang2024qwen2} and Vary~\cite{wei2024small_varytoy}, to further validate the effectiveness of our slow perception.

\subsection{Data Engine}

% The availability of public geometric image data has significantly increased in the community. However, in the field of geometry, datasets providing structured descriptions of geometric diagrams remain scarce. To address this issue, We chose to represent line segments as $(x_1, y_1) -- (x_2, y_2)$, where $(x_1, y_1)$ and $(x_2, y_2)$ are the coordinates of the two endpoints. Circles are represented as $(x_0, y_0, r)$, where $(x_0, y_0)$ is the coordinates of center and $r$ is the radius. This structured information is used as the caption for the geometric diagrams. Specifically, we referred to real geometric diagrams at the middle school level.

We render 200k synthetic geometric images as the train data, wherein Matplotlib is employed as the rendering engine. We stochastically vary multiple parameters to ensure data heterogeneity, including line thickness, line style (solid or dashed), and image resolution (DPI). In total, 150k images are generated with DPI values randomly distributed between 36 and 300, while the remaining 50k are uniformly set to 96 DPI, reflecting a commonly used resolution in practical applications. For the corpus of point-line locations and relationships  that make up geometry, we devise the following generation procedure:

% \begin{itemize}[leftmargin=*]
\noindent\textbf{1) Selection of substrate.} We select the most common quadrilaterals as the rendering base, including squares, rectangles, parallelograms, rhombuses, trapezoids, isosceles trapezoids, right trapezoids, and other uncommon arbitrary quadrilaterals.

\noindent\textbf{2) Addition and deletion of points.} Based on base quadrilaterals, we randomly delete 0-1 points or add 1-6 points to augment the polygon diversity. For extra generated points, we mainly take them on the sides or side extensions of the base figure, with increased probability weights of selecting the mid- or trisection points.

\noindent\textbf{3) Generation circular and text.} With a predetermined probability, we add inscribed and circumscribed circles for the base quadrilaterals. Besides, text labels (``A" to ``Z") are generated at vertex positions with a certain probability. Although these features are not used in slow perception, their inclusion will enhance the resemblance of the rendered data to real-world geometric shapes.

% \end{itemize}
The entire rendering process aforementioned can be represented as follows:
\begin{equation}
\mathbf{G} = \Phi(\Psi(q, \mathbf{P}), \delta, \mathbf{A}, \omega, \rho_c, \rho_t, \mathcal{T})
\label{enq1}
\end{equation}
where $\mathbf{G}$ is the final generated geometric figure; $\Phi$ represents the  geometric figure generation function;
$\Psi$ is base quadrilateral generation function; $q$ $\in$ $\mathcal{Q}$, where Q is the predefined set of quadrilateral types; $\mathbf{P}$=$(p_1, p_2, p_3, p_4)$ represents vectors of initial quadrilateral vertex coordinates; $\delta \in [0, 1]$ is point deletion parameter; $\mathbf{A}$=$(a_1, …, a_n), n \in {0 \ \mathrm{to} \  6}$ is the set of added points; $\omega$ represents the weight factor for special points (e.g., midpoints, trisection points); $\rho_c$ and $\rho_t \in [0, 1]$ are probabilities of generating inscribed/circumscribed circles and text labels; $\mathcal{T}$ is the set of possible text labels.

All rendered points are within the coordinate axis ranging from -10 to 10, with two decimal places retained to ensure accuracy.  During the final label generation, we first use the \textit{TransData} function from Matplotlib to convert the display coordinates to pixel coordinates. Then we recalculate the pixel coordinates to a range between -10 and 10. This is because Matplotlib always auto-adds coordinate axes and padding around the image during rendering, making coordinate conversion difficult. Let $\mathbf{G}$ be a geometric shape as in equation~\ref{enq1}, this process can be represented as:
\begin{equation}
\mathbf{\hat{G}} = \mathrm{Normalize}(\mathrm{TransData}(\mathbf{G})) \times 20 - 10
\label{enq2}
\end{equation}
where $\mathrm{\hat{G}}$ is the geometric shape with final coordinates labels.  Normalization refers to dividing the $x$ and $y$ values of the original coordinates by the width and height of the original image, respectively. This is to unify the coordinate representation of training and testing data, as our validation and test sets are manually annotated.

\begin{figure}[t]
\centering
\includegraphics[width=8.3cm]{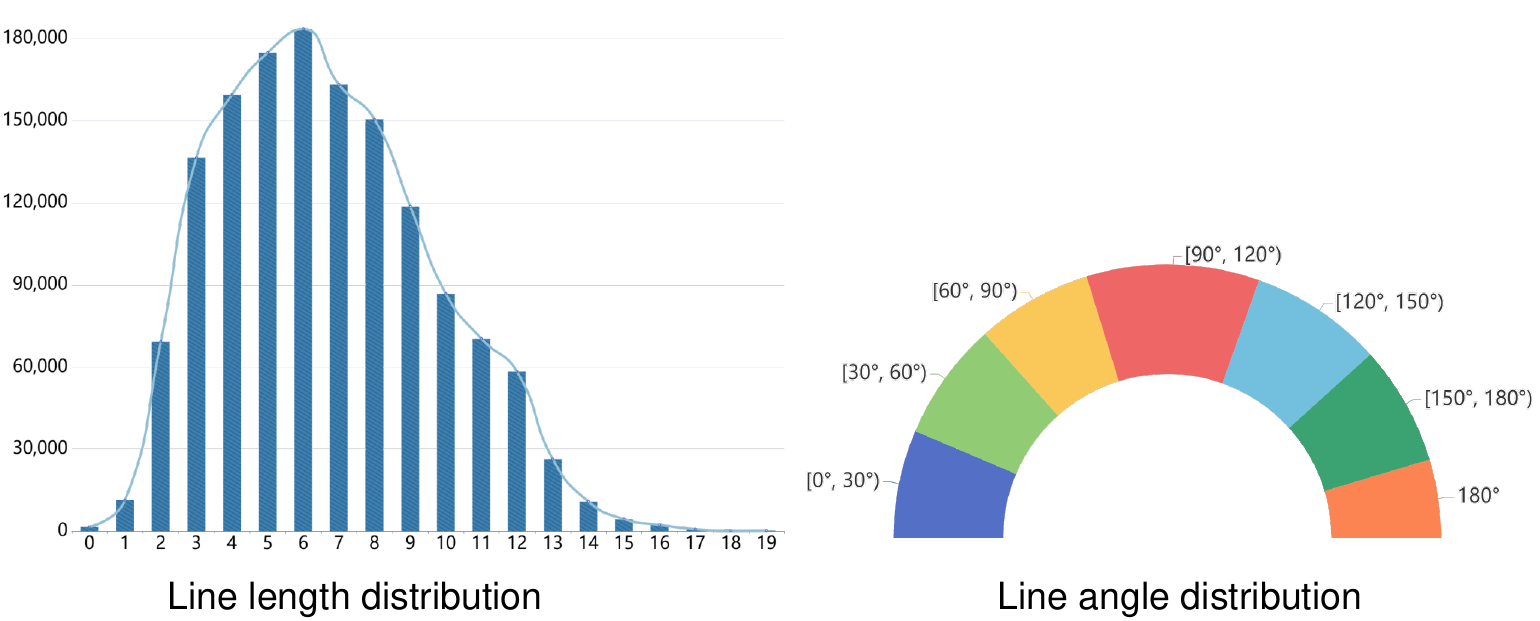}\\
\caption{The line distribution of rendered train data. The left figure shows the line length and the right is angle distributions to comprise the geometric shapes in the train data.}
\label{fig:4} 
\end{figure}

The length and angle distributions of the rendered lines are shown in Figure~\ref{fig:4}. The lengths are mainly distributed between 2 and 10, which will serve as the guidance for setting the perceptual ruler in slow perception.

For evaluation, we use manual manner to construct the benchmark. All images within the benchmark are sourced from the middle school mathematics exam. In total, we collect 480 geometric figures, which are divided into validation and test sets at a ratio of 1:3 via characteristics of samples, resulting in 120 images for the validation and 360 images for the test.

\subsection{Slow Perception}

The proposed slow perception for geometric figure parsing can be mainly divided into two stages: 1) decompose complex geometric figures into basic units and gradually perceive each one. We refer to this process as 1-order slow-down.  2) for each basic point-line pair, we use small local ``perceptual jumps" to slowly and accurately reconstruct it. We name this procedure as 2-order slow-down. The detailed descriptions of them are as follows:

\noindent\textbf{1-order slow down for perception decomposition.}  The main purpose of this stage is to unify the representation form of complex geometric shapes. As shown in the input image of Figure~\ref{Fig3}, there are 8 triangles in total. If take a model to use \textit{Tikz’s} closed shape code \textit{- -cycle} or Matplotlib’s \textit{Polygon} function to draw this shape, issues such as multiple peaks and redundant definitions may easily occur due to multiple triangles are nested within each other. In the 1-order slow-down, we do not need to consider which figure is a polygon. We decompose all figures line-by-line because matter how complex a geometric shape is, it always consists of basic line segments. This ``make complex to simple" process can effectively avoid the multiple peaks problem via a unified representation. Using the above manner, the Figure~\ref{Fig3} input image can be represented as:
\begin{equation}
\mathbf{G'} = 
\begin{cases}
\mathrm{Line}[\ (A,B),(B,C),(C,A),\\
\ \ \ \ \ \ \ \ \ \ (A,D),(A,E),(B,E)]  \\
\mathrm{Circle}[(C_x,C_y, R)]  \\
\end{cases}
\label{enq3}
\end{equation}
where $\mathbf{G'}$ is the geometric shape in Figure~\ref{Fig3}. $\mathrm{Line}$ and $\mathrm{Circle}$ are two sets that contain line units and circle units decoupled from the whole shape.

\noindent\textbf{2-order slow down for perception flow.} In classical computer vision fields, inference time scaling seems to have always existed. For instance, processes such as the transition from proposals to the final bounding box in RPN~\cite{ren2015faster}, or the denoising procedure in diffusion~\cite{ho2020denoising}, is a typical coarse-to-fine inference time scaling manner. However, under the auto-regressive framework, it is not easy for LVLM to perform coarse-to-fine modeling. Therefore, we propose an alternative approach  -- perception flow from local to whole.

Our approach is inspired by sketching techniques. For example, when sketching portraits, there are typically two methods: One involves first constructing the framework and then gradually drawing details. This is called the Contour Method and is a coarse-to-fine approach. The other method starts by drawing details from local areas and slowly builds up to the whole. This is called the Local Method and is a local-to-whole approach.

In geometric parsing tasks, for a long straight line, humans may not draw it accurately in one stroke, and models might face similar challenges. Therefore, we define the maximum single-perception distance (perceptual ruler). Note that for such a 2-order slow-down based on ``multi-stroke flow", we do not apply it to shapes other than lines in this work, due to: a) lines are the most basic and common shapes, and thus need to be prioritized; b) in geometric figures, other shapes, e.g., circle and curve,  have lower interdependencies and will be our future works. 

Let $l$ be the line $\overline{AB} $, wherein the point $A$ is the start point and the $B$ is the end one. $l$ can be redefined via multiple sub-lines  $l_i$:
\begin{equation}
\begin{cases}
l = \overline{AB} = \bigcup_{i=1}^{n} l_i, \quad \text{where} \quad l_i = [x_{i-1}, x_i]; \\ 
x_0 = A, \quad x_n = B, \quad |l_i| = d; \quad \\
\forall i \in \{1, 2, ..., n-1\}, \ \text{and} \quad |l_n| \leq d; \\
n = \lceil |l| / d \rceil
\end{cases}
\label{enq4}
\end{equation}
where $d$ is a hyper-parameter representing the length of the perceptual ruler. $n$ is the number of sub-line segments,  composed of $l$ and $d$ together. Assuming $l = 12$  and $d = 8$, then $n = 2$. If $d = 4$, then $n = 3$. Therefore, when $l$ is fixed, the smaller the perceptual ruler $d$, the more “strokes” are needed to trace a line, resulting in greater computational complexity and increased inference time.
% An example of the ground truth, when $d = 4$, are as follows:

\subsection{Optimization and Evaluation Objectives}

% We 

% \vspace{-2mm}
% \begin{figure}[!h]
% \centering
% \includegraphics[width=3.5cm]{fig/30104.png}\\
% % \caption{The line distribution of rendered train data. The left figure shows the line length and the right is angle distributions to comprise the geometric shapes in train data.}
% % \label{fig:5} 
% \end{figure}

% % \vspace{-5mm}

% % \noindent\textbf{Line:} \\
% (-5.17, 2.33) -- (-2.11, -0.25) -- (0.95, -2.83) -- (3.7, -5.15) \\
% (-5.17, 2.33) -- (-1.73, 4.36) -- (-0.73, 4.95) \\
% (-5.17, 2.33) -- (-1.28, 3.27) -- (2.61, 4.21) -- (5.68, 4.95) \\
% (-0.73, -1.42) -- (-0.73, 2.58) -- (-0.73, 4.95) \\
% (-0.73, -1.42) -- (3.27, -1.42) -- (5.68, -1.42) \\
% (-0.73, 4.95) -- (3.27, 4.95) -- (5.68, 4.95) \\
% (3.7, -5.15) -- (5.58, -1.62) -- (5.68, -1.42) \\
% (5.68, -1.42) -- (5.68, 2.58) -- (5.68, 4.95) \\
% (-0.73, 4.95) -- (0.88, 1.29) -- (2.48, -2.38) -- (3.7, -5.15) \\

\begin{figure}[!t]
\centering
\includegraphics[width=7.5cm]{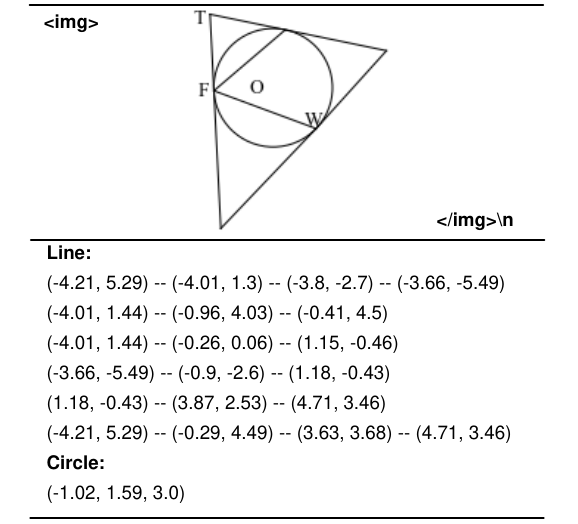}\\
\caption{An example of the ground truth. This figure shows an rendered geometry sample and the corresponding text labels under the length of perceptual ruler being 4.}
\label{fig:5} 
\end{figure}

The input of the model is a geometry image $v$ and the output is the parsing text sequence $t$ . The training optimization objective is as follows:  

\begin{equation}
\mathcal{L}(\omega, t) = -E_{(t, v) \sim D} \log P_\omega\left(t_m \mid t_{<m}, v\right)
\end{equation}
where $w$ denotes the target text sequence, $v$ denotes the vision features from the vision backbone, $m$ denotes the current index of the output target token and $D$ denotes the dataset. $\omega$ represents the model weights. An example of the input image and ground truth text is shown in Figure~\ref{fig:5}. Since we are focusing solely on the geometric figure parsing task, we do not use any prompts.

In evaluation, we use the intersection-over-union (IoU) to determine whether a predicted line segment is a positive or negative sample; specifically, the total IoU of a segment is equal to the average of the horizontal and vertical components. Mathematically,
\begin{equation}
\mathrm{IoU_{\text{line}}} = 
\frac{1}{2} \left( \frac{|P_{\hat{x}} \cap T_{\hat{x}}|}{|P_{\hat{x}} \cup T_{\hat{x}}|} 
+ \frac{|P_{\hat{y}} \cap T_{\hat{y}}|}
{|P_{\hat{y}} \cup T_{\hat{y}}|} \right)
\label{enq6}
\end{equation}
where $P$ is the predicted line segment and $T$ is the ground truth. $\hat{x}$ and $\hat{y}$ denote the the components of the line segment on the $x$-$axis$ and $y$-$axis$, respectively.

\section{Experiment}

\begin{table*}[ht]

	\begin{center}
		\setlength{\tabcolsep}{3mm}{
		
			%\small
			\begin{tabular}{c|c|ccc|ccc|cccc|}
				% \hline  
                    % \toprule
				% \hline
                    \toprule[1pt]
				%	\rowcolor{gray!20}
				Perceptual ruler & IoU & $\mathrm{F1}$  & $\mathrm{F1_{s}}$ & $\mathrm{F1_{l}}$ & $\mathrm{P}$ & $\mathrm{P_{s}}$ & $\mathrm{P_{l}}$ & $\mathrm{R}$ & $\mathrm{R_{s}}$  & $\mathrm{R_{l}}$   \\
				% \hline  % 中部线
                    \midrule
				% \rowcolor{gray!10}
                    \multirow{2}{*}{+$\infty$ (baseline)} & 0.75 & 51.4 & 44.3 & 47.5 & 50.1  & 42.8 & 49.3 & 53.6 & 48.8 & 47.3  \\
                    % \rowcolor{gray!10}
                    \cmidrule{2-11}
                              & 0.9 & 47.5 & 41.6 & 43.7 &  46.3  & 40.1 & 45.2 & 49.5 & 45.9 & 43.6  \\
                    \midrule
                    
                    \multirow{4}{*}{12-length} & \multirow{2}{*}{0.75} & 53.3 & 46.2 & 49.6 & 51.6  & 44.9 & 50.3 & 56.0 & 50.2 & 50.3  \\
                   
                   % \rowcolor{gray!10}
                            &  & \textcolor{red}{$\uparrow 1.9$} & \textcolor{red}{$\uparrow 1.9$} & \textcolor{red}{$\uparrow 2.1$} & \textcolor{red}{$\uparrow 1.5$} & \textcolor{red}{$\uparrow 2.1$} &\textcolor{red}{$\uparrow 1$}&\textcolor{red}{$\uparrow 2.4$} &\textcolor{red}{$\uparrow 1.4$} &\textcolor{red}{$\uparrow 3$}\\
                   % \rowcolor{gray!10}
                   \cmidrule{2-11}
                             & \multirow{2}{*}{0.9} & 49.9 & 43.0 & 47.2 &  48.3  & 41.7 & 47.8 & 52.4 & 46.8 & 47.8  \\ 
                             &  & \textcolor{blue}{$\uparrow 2.4$} & \textcolor{blue}{$\uparrow 1.4$} & \textcolor{blue}{$\uparrow 3.5$} & \textcolor{blue}{$\uparrow 2$} & \textcolor{blue}{$\uparrow 1.6$} &\textcolor{blue}{$\uparrow 2.6$}&\textcolor{blue}{$\uparrow 2.9$} &\textcolor{blue}{$\uparrow 0.9$} &\textcolor{blue}{$\uparrow 4.2$}\\
                    \midrule

                    \multirow{4}{*}{10-length} & \multirow{2}{*}{0.75} & 54.4 & 48.4 & 49.6 & 52.9  & 47.1 & 50.1 & 56.8 & 52.5 & 50.1  \\
                   
                   % \rowcolor{gray!10}
                            &  & \textcolor{red}{$\uparrow 3$} & \textcolor{red}{$\uparrow 4.1$} & \textcolor{red}{$\uparrow 2.1$} & \textcolor{red}{$\uparrow 2.8$} & \textcolor{red}{$\uparrow 4.3$} &\textcolor{red}{$\uparrow 0.8$}&\textcolor{red}{$\uparrow 3.2$} &\textcolor{red}{$\uparrow 3.7$} &\textcolor{red}{$\uparrow 2.8$}\\
                   % \rowcolor{gray!10}
                    \cmidrule{2-11}
                             & \multirow{2}{*}{0.9} & 51.4 & 45.7 & 47.0 &  50.0  & 44.6 & 47.4 & 53.6 & 49.5 & 47.7  \\ 
                             &  & \textcolor{blue}{$\uparrow 3.9$} & \textcolor{blue}{$\uparrow 4.1$} & \textcolor{blue}{$\uparrow 3.3$} & \textcolor{blue}{$\uparrow 3.7$} & \textcolor{blue}{$\uparrow 4.5$} &\textcolor{blue}{$\uparrow 2.2$}&\textcolor{blue}{$\uparrow 4.1$} &\textcolor{blue}{$\uparrow 3.6$} &\textcolor{blue}{$\uparrow 4.1$}\\
                    \midrule

                   \multirow{4}{*}{8-length} & \multirow{2}{*}{0.75} & 55.4 & 50.4 & 49.9 & 54.0  & 49.0 & 51.3 & 57.7 & 54.5 & 49.9  \\
                   
                   % \rowcolor{gray!10}
                            &  & \textcolor{red}{$\uparrow 4$} & \textcolor{red}{$\uparrow 6.1$} & \textcolor{red}{$\uparrow 2.4$} & \textcolor{red}{$\uparrow 3.9$} & \textcolor{red}{$\uparrow 6.2$} &\textcolor{red}{$\uparrow 2$}&\textcolor{red}{$\uparrow 4.1$} &\textcolor{red}{$\uparrow 5.7$} &\textcolor{red}{$\uparrow 2.6$}\\
                   % \rowcolor{gray!10}
                   \cmidrule{2-11}
                             & \multirow{2}{*}{0.9} & 52.1 & 47.3 & 48.0 &  50.7  & 45.9 & 49.3 & 54.3 & 51.1 & 48.0  \\ 
                             &  & \textcolor{blue}{$\uparrow 4.6$} & \textcolor{blue}{$\uparrow 5.7$} & \textcolor{blue}{$\uparrow 4.3$} & \textcolor{blue}{$\uparrow 4.4$} & \textcolor{blue}{$\uparrow 5.8$} &\textcolor{blue}{$\uparrow 4.1$}&\textcolor{blue}{$\uparrow 4.8$} &\textcolor{blue}{$\uparrow 5.2$} &\textcolor{blue}{$\uparrow 4.4$}\\

                    \midrule
                   \multirow{4}{*}{4-length} & \multirow{2}{*}{0.75} & 57.5 & 52.4 & 51.8 & 55.8  & 50.8 & 52.9 & 60.7 & 56.9 & 52.2  \\
                   
                   % \rowcolor{gray!10}
                            &  & \textcolor{red}{$\uparrow 6.1$} & \textcolor{red}{$\uparrow 8.1$} & \textcolor{red}{$\uparrow 4.3$} & \textcolor{red}{$\uparrow 5.7$} & \textcolor{red}{$\uparrow 8$} &\textcolor{red}{$\uparrow 3.6$}&\textcolor{red}{$\uparrow 7.1$} &\textcolor{red}{$\uparrow 8.1$} &\textcolor{red}{$\uparrow 4.9$}\\
                   % \rowcolor{gray!10}
                   \cmidrule{2-11}
                             & \multirow{2}{*}{0.9} & 53.5 & 47.3 & 49.5 &  51.9  & 45.9 & 50.4 & 56.0 & 51.2 & 49.9  \\ 
                             &  & \textcolor{blue}{$\uparrow 6$} & \textcolor{blue}{$\uparrow 5.7$} & \textcolor{blue}{$\uparrow 5.8$} & \textcolor{blue}{$\uparrow 5.6$} & \textcolor{blue}{$\uparrow 5.8$} &\textcolor{blue}{$\uparrow 5.2$}&\textcolor{blue}{$\uparrow 6.5$} &\textcolor{blue}{$\uparrow 5.3$} &\textcolor{blue}{$\uparrow 6.3$}\\

        % \hline
        \bottomrule[1pt]
		\end{tabular}}

	\end{center}
	\caption{Results of different manners on the SP-1 \textbf{test-set}. Here, “s” and “l” are abbreviations for “short” and “long,” representing short segments and long segments, respectively. The threshold is set at 8, with segments less than 8 considered as short and those greater than 8 as long. The red upward arrow \textcolor{red}{$\uparrow$} indicates the improvement of the current method over the baseline at 0.75 IoU, while the blue ones \textcolor{blue}{$\uparrow$} signifies the performance improvement under 0.9 IoU.}
	\label{table1}
    \end{table*}

\begin{table*}[ht]

	\begin{center}
		\setlength{\tabcolsep}{3mm}{
		
			%\small
			\begin{tabular}{c|c|ccc|ccc|cccc|}
				% \hline  
                    % \toprule
				% \hline
                    \toprule[1pt]
				%	\rowcolor{gray!20}
				Perceptual ruler & IoU & $\mathrm{F1}$  & $\mathrm{F1_{s}}$ & $\mathrm{F1_{l}}$ & $\mathrm{P}$ & $\mathrm{P_{s}}$ & $\mathrm{P_{l}}$ & $\mathrm{R}$ & $\mathrm{R_{s}}$  & $\mathrm{R_{l}}$   \\
				% \hline  % 中部线
                    \midrule
				% \rowcolor{gray!10}
                    \multirow{2}{*}{+$\infty$ (baseline)} & 0.75 & 52.2 & 41.3 & 49.2 & 51.1  & 39.2 & 50.6 & 53.7 & 46.6 & 48.9  \\
                    % \rowcolor{gray!10}
                    \cmidrule{2-11}
                              & 0.9 & 48.6 & 36.4 & 47.2 &  47.6  & 34.9 & 48.6 & 50.1 & 40.6 & 46.8  \\

                    \midrule
                   \multirow{4}{*}{4-length} & \multirow{2}{*}{0.75} & 56.7 & 44.3 & 54.3 & 54.9  & 42.0 & 55.5 & 59.5 & 49.6 & 54.4  \\
                   
                   % \rowcolor{gray!10}
                            &  & \textcolor{red}{$\uparrow 4.5$} & \textcolor{red}{$\uparrow 3$} & \textcolor{red}{$\uparrow 5.1$} & \textcolor{red}{$\uparrow 3.8$} & \textcolor{red}{$\uparrow 2.8$} &\textcolor{red}{$\uparrow 4.9$}&\textcolor{red}{$\uparrow 5.8$} &\textcolor{red}{$\uparrow 3$} &\textcolor{red}{$\uparrow 5.5$}\\
                   % \rowcolor{gray!10}
                   \cmidrule{2-11}
                             & \multirow{2}{*}{0.9} & 51.9 & 39.0 & 51.6 &  50.3  & 37.2 & 52.8 & 54.2 & 43.1 & 51.6  \\ 
                             &  & \textcolor{blue}{$\uparrow 3.3$} & \textcolor{blue}{$\uparrow 2.6$} & \textcolor{blue}{$\uparrow 4.4$} & \textcolor{blue}{$\uparrow 2.7$} & \textcolor{blue}{$\uparrow 2.3$} &\textcolor{blue}{$\uparrow 4.2$}&\textcolor{blue}{$\uparrow 4.1$} &\textcolor{blue}{$\uparrow 2.5$} &\textcolor{blue}{$\uparrow 4.8$}\\

        % \hline
        \bottomrule[1pt]
		\end{tabular}}

	\end{center}
	\caption{Results of different manners on the SP-1 \textbf{val-set}. The up-arrow in the figure has the same meaning as Table~\ref{table1}. It can be seen that the performance improvements of slow perception on the validation split are also stable.}
	\label{table2}
    \end{table*}

\subsection{Experimental Settings}

\noindent\textbf{Datasets:} We name the train data and benchmarks we generated as SP-1, including 200k synthetic image-text pairs for trianing and 480 real-scenario samples for evaluation.  We divide the evaluation part into a validation set and a test set, with a ratio of 1:3, resulting in 120 images for validation and 360 images for test. All the data we used will be open-sourced to promote the advancement of geometric figure parsing. We also hope that our data configuration will serve as the de-facto setup for subsequent followers to ensure a fair comparison.

\noindent \textbf{Implementation details.}  We select three models for experiments: GOT~\cite{wei2024general}, Qwen2-VL-2B~\cite{wang2024qwen2}, and Vary-toy~\cite{wei2024small_varytoy}. GOT is the primary model for slow perception, and we conduct most of our experiments on it because it offers good performance, has a smaller model size, and allows for fast iteration. Qwen2-VL and Vary-toy serve as auxiliary models to provide more solid evidence for our conclusions. For GOT, we unfreeze all parameters for training. For Qwen2-VL and Vary-toy, we freeze the encoder parameters and unfreeze the LLM part for fine-tuning. All other experimental settings are identical. Specifically, we use 8 L40s GPUs for training, run 2 epochs on the SP-1 dataset, with a per-GPU batch size of 2 and a gradient accumulation of 2, resulting in a global batch size of 32. Simple data augmentations, e.g., color/lighting jitter and Gaussian noise, are utilized. We employ cosine annealing~\cite{loshchilov2016sgdr} to adjust the learning rate, starting at 3e-5, with a total of 12,500 iterations and a warm-up ration of 0.003. Training GOT takes about 3 hours, Vary-toy needs 5 hours, and Qwen2-VL-2B, due to its larger encoder~\cite{radford2021learning} computational cost, requires 15 hours.

\noindent \textbf{Baseline definition.}  We do not define model-level baselines because the direct testing performance of all models on our val/test set are too low. Instead, our baseline is method-level defined.  Specifically, by training SP-1, we set a perception ruler of infinite length as the baseline, meaning that for each line, a baseline model always directly regress from the starting point to the endpoint without slow perception.

% \subsection{Evaluation Metrics}

\noindent \textbf{Evaluation metrics.} We use F1-score, precision, and recall to measure the effectiveness of different methods. Specifically, we utilize the IoU from equation~\ref{enq6} to determine whether a prediction is a positive or negative sample. The basic IoU threshold is 0.75, and the strict threshold is 0.9. Precision, recall, and  F1-score are defined as: $\text{P}$=$\frac{\mathrm{TP}}{(\mathrm{TP}+\mathrm{FP})}$, $\text{R}$=$\frac{\mathrm{TP}}{(\mathrm{TP}+\mathrm{FN})}$,
% \begin{equation}
% \text { Precision }=\frac{\mathrm{TP}}{(\mathrm{TP}+\mathrm{FP})}
% \end{equation}
% \begin{equation}
% \text { Recall }=\frac{\mathrm{TP}}{(\mathrm{TP}+\mathrm{FN})}
% \end{equation}
where $\mathrm{TP}$, $\mathrm{FP}$ and  $\mathrm{FN}$ represent true positives, false positives, and false negatives, respectively. With the calculated precision and recall values, the F1-score can be further computed as: F1 = $2$$\times$(\text{P}$\times$\text{R})$/$({\text{P}+\text{R}}). F1-score is generally considered to be a balance between precision and recall, and it is our main metric for measuring the proposed slow perception performance.

\subsection{Main Results}

\begin{figure}[t]
\centering
\includegraphics[width=7.9cm]{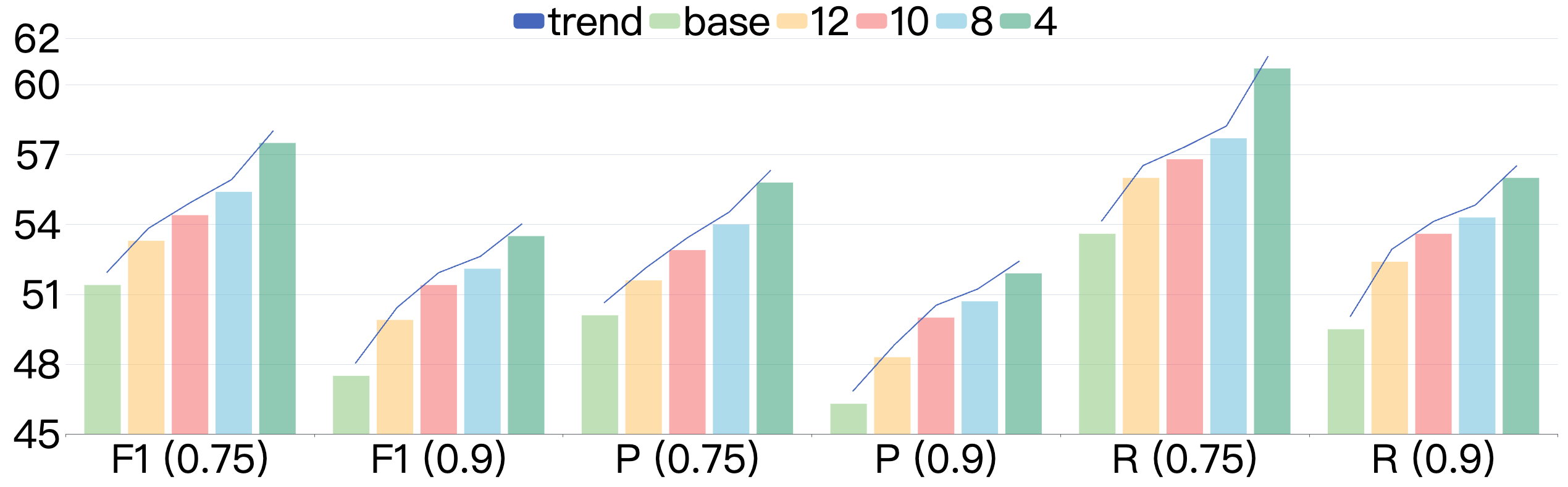}\\
\caption{As the length of the perceptual ruler decreases, we can observe a steady improvement in almost all metrics. The shorter the perceptual ruler, the more “strokes” are needed to model a line, resulting in the model outputting more intermediate “gaze” points. This leads to increased computational complexity during inference, and correspondingly longer inference times, exhibiting to some extent an inference time scaling law.}
\label{fig:6} 
\end{figure}

\noindent\textbf{Effectiveness of slow perception.} Table~\ref{table1} shows the performance comparison of slow perception on the SP-1 test set. All results are obtained from training the GOT-OCR2.0~\cite{wei2024general} model. The baseline (prediction line segment from the starting point to the ending point) can achieve an F1-score of 51.4\%, a precision of 50.1\%, and a recall of 53.6\% at 0.75 IoU. When the criterion becomes stricter (0.9 IoU), these values decrease to 47.5\%, 46.3\%, and 49.5\%, respectively. As the slow perception method is introduced, performance gradually improves. It can be observed that when using a relatively long perceptual ruler (12-length), the F1-score can be increased by 1.9\%, precision by 1.5\%, and recall by 2.4\% at 0.75 IoU. The improvement becomes even more pronounced at 0.9 IoU.  As the perceptual ruler length gradually decreases from 12 to 4, we can observe an almost steady increase in performance metrics. At a length of 4, the slow perception method outperforms the baseline by 6.1\% in F1-score, 5.7\% in precision, and 7.1\% in recall at 0.75 IoU. These results strongly demonstrate the effectiveness of slow perception.

For images in the validation set, apart from having a different sample size compared to the test set, the geometric shapes show more complex interweaving of short lines, meaning that short lines are more challenging to predict while long lines are slightly easier than those in the test set. As shown in Table~\ref{table2}, although the improvements from slow perception are lower than those in the test set, the lifts remain substantial. Slow perception mainly solves the problem of long line segments in a ``single stroke" of a model through ``perception flow" and results on val-set aligns with this feature, which further corroborates that the effect of the proposed slow perception is solid.

\noindent\textbf{Inference time scaling law.} Figure~\ref{fig:6} presents a visual chart of the slow perception performance on the test set, which clearly demonstrates an inference time scaling law - longer inference times correlate with better model performance. This may be due to the model having an upper limit on its precise perception distance, similar to human perception. We believe that this perceptual inference scaling could also provide insights for other computer vision tasks.

\begin{table}[ht]

	\begin{center}
		\setlength{\tabcolsep}{3mm}{
		
			%\small
			\begin{tabular}{l|c|c|ccc}
				% \hline  
                    % \toprule
				% \hline
                    \toprule[1pt]
				%	\rowcolor{gray!20}
				Model & Size & Ruler & $\mathrm{F1}$  & $\mathrm{P}$ & $\mathrm{R}$   \\
				% \hline  % 中部线
                    \midrule
				% \rowcolor{gray!10}
                    \multirow{2}{*}{Qwen2-VL} &\multirow{2}{*}{2B} & +$\infty$ & 44.1 & 43.1 & 46.0   \\

                              & & 4 & 46.0 & 45.2 & 47.9   \\
                    
                    \midrule
                    \multirow{2}{*}{Vary-toy} &\multirow{2}{*}{1.8B} & +$\infty$ & 45.5 & 44.8 & 47.2   \\
                             &  & 4 & 47.8 & 46.7 & 50.0   \\
                    % \midrule
                   % \multirow{2}{*}{4-length} & \multirow{2}{*}{0.75} & 56.7 & 44.3 & 54.3 & 54.9   \\
                   % & 0.9 & 48.6 & 36.4 & 47.2 
                    
        % \hline
        \bottomrule[1pt]
		\end{tabular}}

	\end{center}
	\caption{Slow perception on other LVLMs. We freeze the encoders to train Qwen2-VL~\cite{wang2024qwen2} and Vary-toy~\cite{wei2023vary} and test these models on SP-1 test-set to further verify the efficiency of the proposed method. The ``Ruler" means the perceptual ruler length, and thus +$\infty$ represents the baseline without slow perception.}
	\label{table3}
    \end{table}

\begin{table}[ht]

	\begin{center}
		\setlength{\tabcolsep}{2.85mm}{
		
			%\small
			\begin{tabular}{l|c|c|ccc}
				% \hline  
                    % \toprule
				% \hline
                    \toprule[1pt]
				%	\rowcolor{gray!20}
				Model & Unfreeze & Ruler & $\mathrm{F1}$  & $\mathrm{P}$ & $\mathrm{R}$   \\
				% \hline  % 中部线
                    \midrule
				% \rowcolor{gray!10}
                    \multirow{4}{*}{GOT} & $\checkmark$ & +$\infty$ & 51.4 & 50.1 & 53.6   \\

                              & $\checkmark$ & 4 & 57.5 & 55.8 & 60.7   \\
                    \cmidrule{2-6}
                    % \midrule
                     & $\times$ & +$\infty$ & 43.8 & 41.7 & 47.3   \\
                             & $\times$ & 4 & 46.9 & 44.2 & 50.9   \\
                    % \midrule
                   % \multirow{2}{*}{4-length} & \multirow{2}{*}{0.75} & 56.7 & 44.3 & 54.3 & 54.9   \\
                   % & 0.9 & 48.6 & 36.4 & 47.2 
                    
        % \hline
        \bottomrule[1pt]
		\end{tabular}}

	\end{center}
	\caption{Vision encoder test. We further test whether the vision encoder is a bottleneck for geometric figure parsing task by freezing or unfreezing the GOT~\cite{wei2024general} encoder.}
	\label{table4}
    \end{table}
    
\subsection{Ablation Study}

\noindent\textbf{Slow perception on other LVLMs.} The above experiments are conducted based on the GOT~\cite{wei2024general} model. To verify the stability of the proposed slow perception, we select two other LVLMs for training and testing, i.e., Qwen2-VL~\cite{wang2024qwen2} and Vary-toy~\cite{wei2024small_varytoy}. Both have decoders of around 2B parameters, and we freeze their encoders during training to save GPU resources. As shown in Table~\ref{table3}, both models perform much lower than that of GOT in Table~\ref{table1} (with the encoder unfrozen). We think the bottleneck may lie in their original (CLIP~\cite{radford2021learning}) encoders’ insufficient ability to perceive geometric points and lines. Even so, slow perception still achieve a stable about 2\% performance increase, which fully demonstrates its robustness.

\noindent\textbf{Vision encoder bottleneck.} Table~\ref{table4} shows the test results after training GOT by freezing and unfreezing its vision encoder. It can be seen that unfreezing the encoder significantly improves the performance of baseline, and after unfreezing, the improvement from slow perception is even greater. This suggests that there is still considerable room in the research and training of encoders in current LVLMs.

\begin{table}[ht]

	\begin{center}
		\setlength{\tabcolsep}{2.85mm}{
		
			%\small
			\begin{tabular}{l|c|c|ccc}
				% \hline  
                    % \toprule
				% \hline
                    \toprule[1pt]
				%	\rowcolor{gray!20}
				Model & Jitter & Ruler & $\mathrm{F1}$  & $\mathrm{P}$ & $\mathrm{R}$   \\
				% \hline  % 中部线
                    \midrule
				% \rowcolor{gray!10}
                    \multirow{2}{*}{GOT} & $\times$ & 4 & 57.5 & 55.8 & 60.7   \\

                              & $\checkmark$ & 4 & 56.6 & 54.5 & 59.6   \\
                    % \cmidrule{2-6}
                    % % \midrule
                    %  & $\times$ & +$\infty$ & 43.8 & 41.7 & 47.3   \\
                    %          & $\times$ & 4 & 46.9 & 44.2 & 50.9   \\
                    % \midrule
                   % \multirow{2}{*}{4-length} & \multirow{2}{*}{0.75} & 56.7 & 44.3 & 54.3 & 54.9   \\
                   % & 0.9 & 48.6 & 36.4 & 47.2 
                    
        % \hline
        \bottomrule[1pt]
		\end{tabular}}

	\end{center}
	\caption{Which is more important, the accuracy of the gaze point or the perception flow? We randomly jitter the ground truth of ``gaze points'' along the line segment. The performance only decrease by less than 1\% (57.5\% vs. 56.6\%).}
	\label{table5}
    \end{table}

\vspace{-5mm}
\begin{figure}[h]
\centering
\includegraphics[width=8cm]{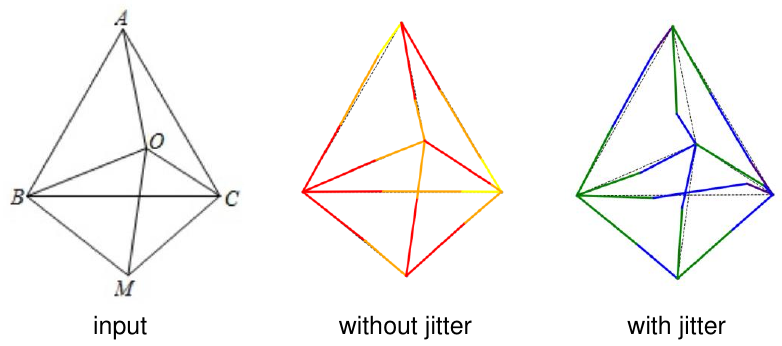}\\
\caption{ `With jitter" represents the result of a trained model using gaze points that have been shaken. The ``stroke order" of each line segment is mapped according to the color of the rainbow, e.g., red, orange, and yellow are used in ``without jitter" result, and green, cyan, and blue are used in ``with jitter" one.}
\label{fig:7} 
\end{figure}

\noindent\textbf{Gaze points jitter. } We randomly jitter the ground truth of the additional ``gaze points" along line segments, with jittering ranges from 0 to 1/10 of the line segment length, to test which is more important in 2-order slow-down of slow perception: accurate prediction of gaze points or the flow of perception. From the Table~\ref{table5}, we can observe that adding noise to gaze points only affects performance by less than 1\%. Even with imprecise gaze points, the model’s performance under slow perception remains far superior to the baseline (56.6\% vs. 51.4\% on F1-score). This suggests that the  perception flow procedure, i.e., the process of gradually perceiving from the start-point to the end-point, may be the core of slow perception. As shown in Figure~\ref{fig:7}, the accuracy of intermediate process of perceptual flow has minimal impact on the final endpoints. This conclusion mitigates the difficulty of gaze point annotation and may inspire us to extend slow perception to more general scenarios.

\begin{figure}[t]
\centering
\includegraphics[width=8.35cm]{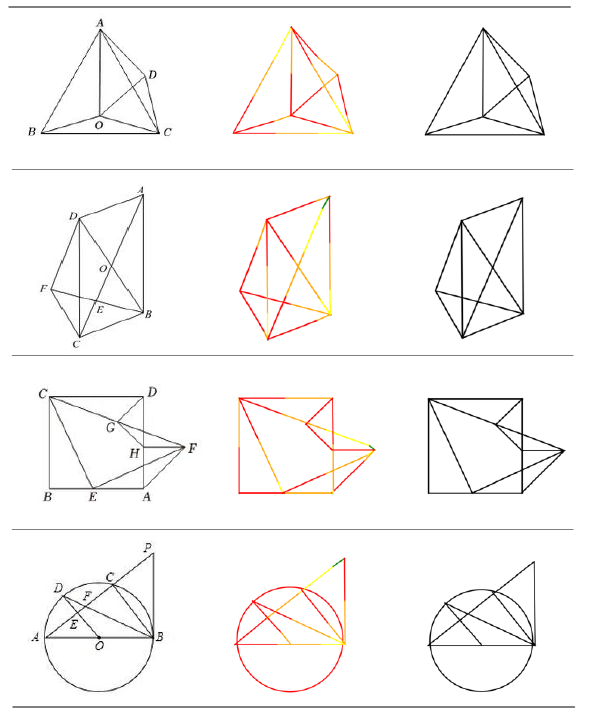}\\
\caption{Slow perception visualization results. The first column represents the input image, and the second column shows the trace route of each ``stroke" executed by the model in slow perception, with ``stroke order" defined by rainbow colors. The third column is the final result of parsing slow perception.}
\label{fig:8} 
\end{figure}

\subsection{Visualization Result}
We provide visualization results to better understand the operation of slow perception. As shown in Figure~\ref{fig:8}, based on slow perception, the model gradually draws from the start-point to the ending when drawing each line segment by multiple ``strokes". This process seems including a gradual correction process in a human-like strategy.
\section{Conclusion}

By consensus, ``visual o1" is a promising direction and a necessary step towards AGI. However, the community seems to skip the most fundamental perception and trend to make LVLMs directly solve visual reasoning problems, e.g., mathematics in geometry. We argue that solving perception is the first step of visual o1; if a model can’t even accurately copy visual geometry, how can we expect it to directly answer complex reasoning questions correctly? In this paper, we propose slow perception for geometric parsing tasks and our results show it is very effective. A geometric figure is an abstraction of natural visual scenes by humans, and we believe that our slow perception approach can also be inspiring for other general vision areas.

% \newpage

\section{Appendix}

In the main text, we primarily discuss the value of slow perception in the research field, focusing on how fine-grained perception tasks require the decomposition and flow of perception. This appendix section will further demonstrate the usage skills of slow perception in downstream application scenarios.  

Because in real practical scenarios, there is a gap between the geometric images and those that we rendered for training. Therefore, we add some of the in-house real data for post-training. Note that this is only to further show our exploration of geometric parsing based on slow perception and does not affect all conclusions of slow perception in the main body.

\begin{figure}[h]
\centering
\includegraphics[width=8.2cm]{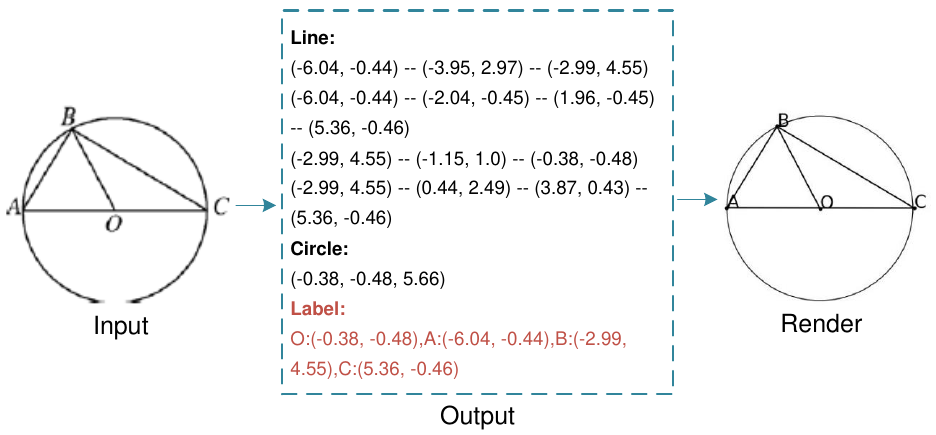}\\
\caption{Adding labels for points and lines in geometric shapes is easy for the auto-regression framework. Although this process does not affect the claim of slow perception, it is necessary to embed the geometry parsing results into downstream tasks, e.g., mathematic geometric VQA. }
\label{fig:9} 
\end{figure}

\subsection{More Complete Geometric Shapes}
In geometric parsing applications, in addition to the coordinates and relationships of point-lines, sometimes we also need labels for them to support downstream business. The task itself is not related to slow perception, but since our method is based on the LVLM~\cite{wei2024general} framework, implementing this feature is very simple, i.e., you only need to simply add key-value pairs corresponding to labels in ground-truth to train the model, as shown in Figure~\ref{fig:9}.

\subsection{From Geometric Parsing to Reasoning}

\begin{figure}[h]
\centering
\includegraphics[width=7cm]{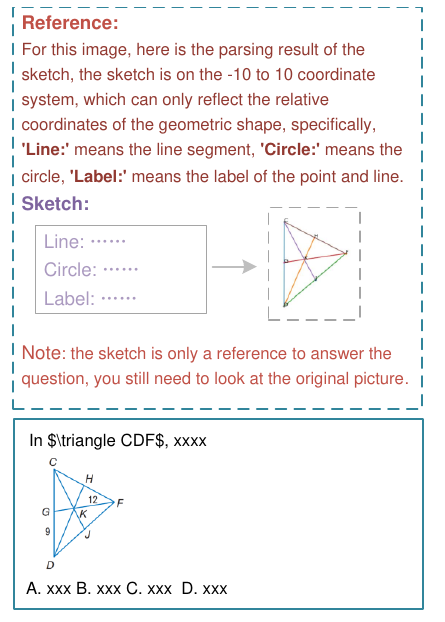}\\
\caption{The organizational of input when adding geometry parsing results as a reference for GPT-4o. We provide the parsing results as a ``sketch" to GPT-4o, emphasizing that it can only represent the relationship between points and lines to a certain extent, and is only for reference.  We require the model that the final answer still needs to be based on the input image.}
\label{fig:10} 
\end{figure}

We use Mathvista~\cite{lu2023mathvista} Geo-subset to further verify the efficiency of geometric parsing based on slow perception for LVLMs on question-answer tasks. The Geo-subset includes 208 images.  We select the state-of-the-art LVLM, GPT-4o, as the experiment target and utilize the 4-ruler slow perception GOT~\cite{wei2024general} with reality data post-training to generate a parsing reference.  With the parsing results, we organize the additional reference to GPT-4o as Figure~\ref{fig:10}.

\begin{table}[ht]

	\begin{center}
		\setlength{\tabcolsep}{2.85mm}{
		
			%\small
			\begin{tabular}{l|c|c}
				% \hline  
                    % \toprule
				% \hline
                    \toprule[1pt]
				%	\rowcolor{gray!20}
				Model & Method &Accuracy    \\
				% \hline  % 中部线
                    \midrule
				% \rowcolor{gray!10}
                    \multirow{2}{*}{GPT-4o} & original & 53.37   \\

                              & + slow perception & 60.10 \textcolor{red}{$\uparrow 6.73$}  \\

        % \hline
        \bottomrule[1pt]
		\end{tabular}}

	\end{center}
	\caption{With geometric parsing results as a reference. GPT-4o can lift 6.73\% accuracy on the Mathvista geo subset. This result further indicates that even for GPT-4o, its fine-grained visual perception ability is insufficient, perception is the foundation of reasoning, and its difficulty has always been overlooked.}
	\label{table6}
    \end{table}

% \vspace{-2mm}

\begin{figure*}[!t]%[h!]
	\centering
	\includegraphics[width=17.2cm]{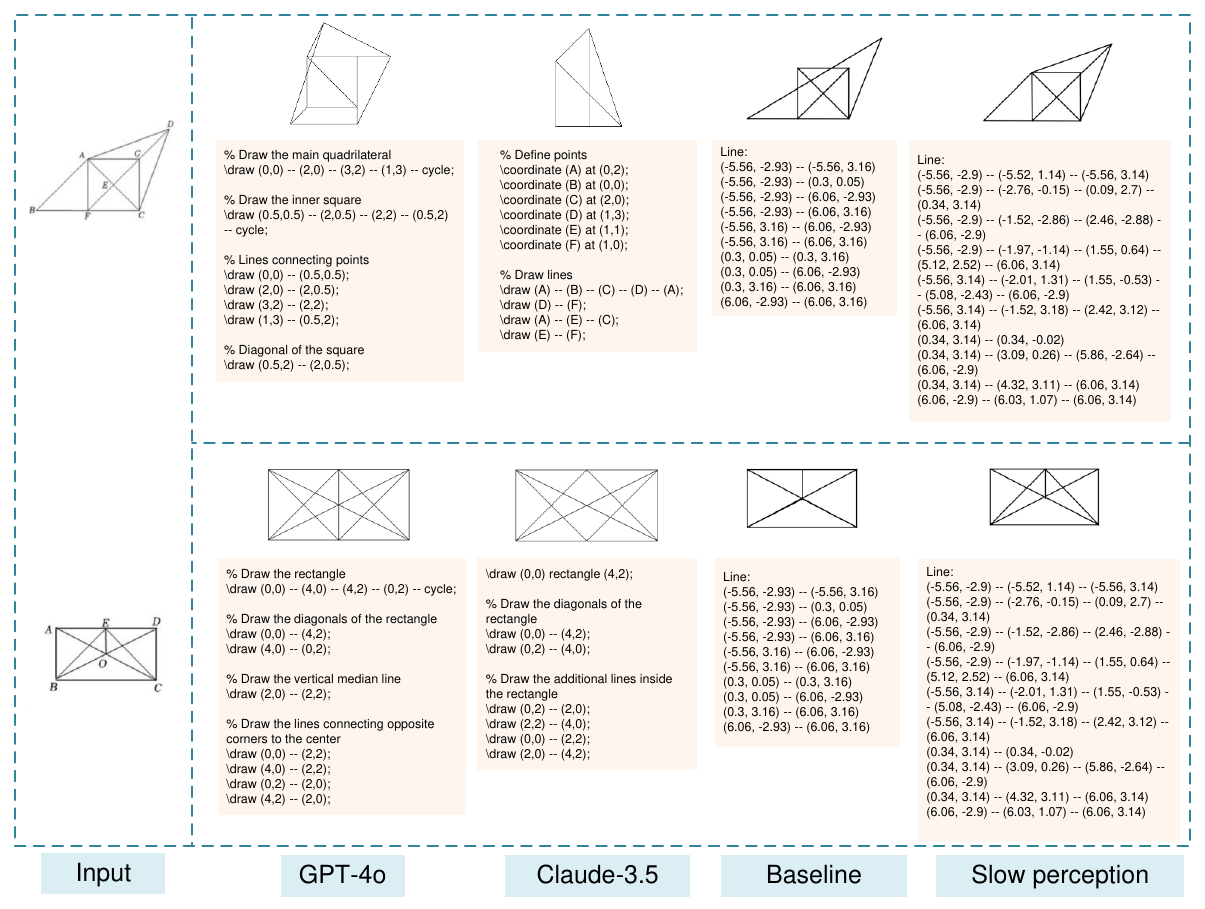}
	%\caption{pic1}
	%\vspace{-2mm}
	\caption{Visualization of geometric parsing results of different models. For GPT-4o and Claude-3.5, we use this prompt to output the results: \textit{Write the Tikz code for this geometric figure, be careful not to write labels for points, only draw the geometric shape}.}
	\label{fig11}
\end{figure*}

As shown in Table~\ref{table6}, for original results without parsing reference, GPT-4o can achieve 53.37\% accuracy. When we add the parsing results as a reference, the accuracy lifts to  60.10\%. This experimental result proves that LVLMs, even GPT-4o, suffer obvious shortcomings in perception, and the community overlooks perception. We believe the slow perception concept, specifically the perception inference time scaling, may be a nice solution.

However, parsing geometric figures beforehand and then using texts to help the model is not the optimal way. A more human-like approach would be for the model to learn to repeatedly look at the image during problem-solving and to draw relevant auxiliary lines at the appropriate times. This depends on the model being able to read images more naturally, wherein the ``perception o1" is a key. 

% and GPT4o to test the efficiency of geometric parsing based on slow perception to geometric figure question-answer problems.

\subsection{Visualization Results Comparison}

Figure~\ref{fig11} shows the visual comparison of slow perception, baseline, and two other advanced LVLMs, i.e., GPT-4o~\cite{GPT4} and Claude-3.5~\cite{AnthropicClaude} in geometric parsing tasks. We utilize the prompt \textit{``Write the Tikz code for this geometric figure, be careful not to write labels for points, only draw the geometric shape"} to make GPT-4o and Claude-3.5 output Tikz code and use the \LaTeX ~to render the results. It can be seen that the two most advanced models can not output satisfactory results on the geometric fine-grained parsing, and such a task may be much more difficult than expected. Different from the output Tikz code, the baseline model uses the 1-order slow-down data (which can be understood as half of the slow perception) proposed in the paper, which splits the geometric shape more atomically. The results show that its output is closer to the input, but it is prone to wrong lines.  The model output using all slow perception methods is better, which shows that the modeling method of slow perception is more reasonable for the optimization of line segments.

\subsection{Future Outlook}

Geometric parsing is just an entry point for slow perception. Essentially, we aim to find a reasonable method to increase the computational complexity of perceptual task reasoning. This method should meet the following requirements: the computational complexity should vary according to the difficulty of different targets that the model perceives, such as long lines versus short lines in this paper. If this can be extended to general scenarios in the future, it would be analogous to occluded versus non-occluded objects in object detection. 

Moving forward, we will focus on two aspects. First, we plan to introduce reinforcement learning to make slow perception more elegant in geometric parsing tasks, akin to a variable-length perceptual ruler. Second, we aim to apply this idea to more generalized tasks.

{\small
\bibliographystyle{ieee_fullname}
\bibliography{egbib}
}

\end{document}